\documentclass[lettersize,journal]{IEEEtran}
\usepackage{amsmath,amsfonts}
\usepackage{algorithmic}
\usepackage{algorithm}
\usepackage{array}
\usepackage[caption=false,font=normalsize,labelfont=sf,textfont=sf]{subfig}
\usepackage{textcomp}
\usepackage{stfloats}
\usepackage{url}
\usepackage{verbatim}
\usepackage{graphicx}
\usepackage{xcolor}
\usepackage{cite}
\usepackage{xcolor}
\usepackage{booktabs}
\usepackage{amssymb}
\usepackage{multirow}
\usepackage{pifont}
\usepackage{url}
\usepackage{hyperref}
\usepackage{tabularx}
\usepackage{multirow}
\usepackage{bbding}
\usepackage{amssymb}
\usepackage{amsmath}
\usepackage{graphicx}
\usepackage{subcaption}
\usepackage{textcomp}
\usepackage{stfloats}
\usepackage{url}
\usepackage{verbatim}
\usepackage{graphicx}
\usepackage{cite}
\usepackage{booktabs}
\usepackage{amssymb}
\usepackage{multirow}
\usepackage{pifont}
\usepackage{url}
\usepackage{hyperref}
\usepackage{tabularx}
\usepackage{multirow}
\usepackage{bbding}

\usepackage{cite}
\usepackage{amsmath,amssymb,amsfonts}
\usepackage{algorithmic}
\usepackage{graphicx}
\usepackage{textcomp}
\usepackage{bm}

\def\BibTeX{{\rm B\kern-.05em{\sc i\kern-.025em b}\kern-.08em
    T\kern-.1667em\lower.7ex\hbox{E}\kern-.125emX}}

\begin{document}

\title{Improving Weakly-supervised Video Instance Segmentation by Leveraging Spatio-temporal Consistency \\}

\author{Farnoosh Arefi, Amir M. Mansourian, Shohreh Kasaei
\thanks{The authors are with the Department of Computer Engineering, Sharif
University of Technology, Tehran 11155, Iran (e-mail: far.arefi@sharif.edu; amir.mansurian@sharif.edu; kasaei@sharif.edu).}
}

\maketitle

\begin{abstract}
The performance of Video Instance Segmentation (VIS) methods has improved significantly with the advent of transformer networks. However, these networks often face challenges in training due to the high annotation cost. To address this, unsupervised and weakly-supervised methods have been developed to reduce the dependency on annotations. This work introduces a novel weakly-supervised method called Eigen-Cluster VIS that, without requiring any mask annotations, achieves competitive accuracy compared to other VIS approaches. This method is based on two key innovations: a Temporal Eigenvalue Loss (TEL) and a clip-level Quality Cluster Coefficient (QCC). The TEL ensures temporal coherence by leveraging the eigenvalues of the Laplacian matrix derived from graph adjacency matrices. By minimizing the mean absolute error between the eigenvalues of adjacent frames, this loss function promotes smooth transitions and stable segmentation boundaries over time, reducing temporal discontinuities and improving overall segmentation quality. The QCC employs the K-means method to ensure the quality of spatio-temporal clusters without relying on ground truth masks. Using the Davies-Bouldin score, the QCC provides an unsupervised measure of feature discrimination, allowing the model to self-evaluate and adapt to varying object distributions, enhancing robustness during the testing phase. These enhancements are computationally efficient and straightforward, offering significant performance gains without additional annotated data. The proposed Eigen-Cluster VIS method is evaluated on the YouTube-Video Instance Segmentation (YouTube-VIS) 2019/2021 and Occluded Video Instance Segmentation (OVIS) datasets, demonstrating that it effectively narrows the performance gap between the fully-supervised and weakly-supervised VIS approaches. The code is available on: \textnormal{\href{https://github.com/farnooshar/EigenClusterVIS}{https://github.com/farnooshar/EigenClusterVIS}}
\end{abstract}

\begin{IEEEkeywords}
Video instance segmentation, Weakly-supervised learning, Deep spectral methods
\end{IEEEkeywords}

\section{Introduction}

\IEEEPARstart{O}bject tracking and segmentation are two fundamental tasks in the field of computer vision with vast applications in areas like autonomous driving \cite{voigtlaender2019mots}, surveillance \cite{elhoseny2020multi}, and sports analysis \cite{mansourian2023multi, somers2024soccernet}. Video Instance Segmentation \cite{yang2019video}, on the other hand, involves both segmenting and tracking instances across all video frames. Despite its challenging nature, there has been rapid progress in deep learning-based methods for this task, thanks to the availability of annotated datasets for the VIS \cite{yang2019video, qi2022occluded}. Although fully supervised methods have achieved considerable performance on these datasets, annotating many videos for this task is heavily labor-intensive and limits the dataset to a limited number of domains and classes. Specifically, this is an important problem for transformer-based VIS models \cite{wang2021end, wu2022seqformer, cheng2021mask2former}, which are supposed to be very data-hungry to achieve their best performance.


\begin{figure}[!ht]
        \raggedleft
	\centering
	\centerline{\includegraphics[scale=0.22]{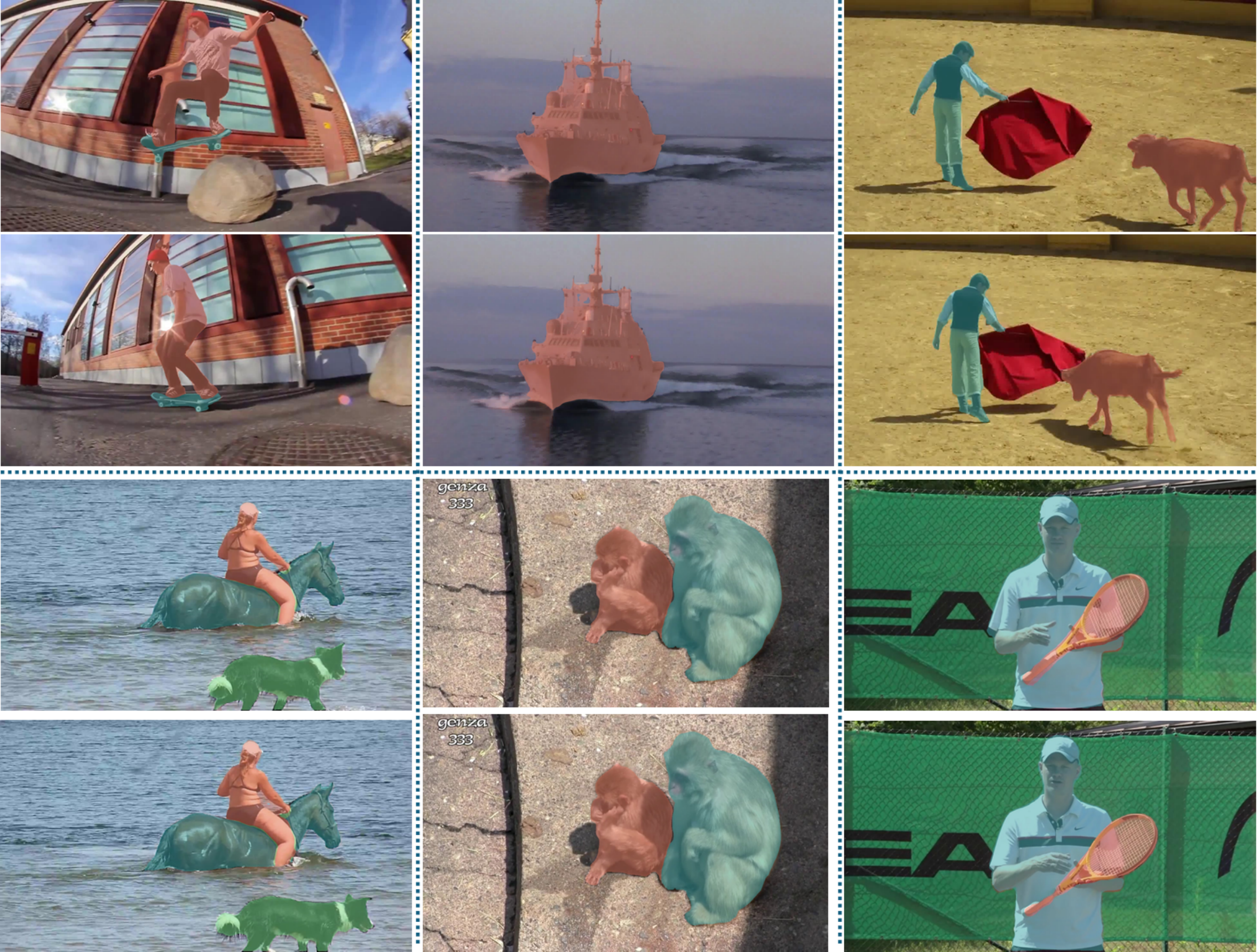}}
	\caption{{\bf Visualization of Eigen-Cluster VIS Predictions. Eigen-Cluster VIS uses no video or image mask annotations, and achieves 44.2\% mask AP on the YouTube-VIS 2019 validation dataset, with a ResNet-50 backbone. These outputs validate the fact that high performance VIS can be obtained without any mask annotations.}}
	\label{fig: pull_figure}
\end{figure}

To address the aforementioned challenges, a various weakly-supervised \cite{khoreva2017simple, lee2021bbam, cheng2022pointly, lee2024extreme, fothi2024cluster2former} and unsupervised\cite{wang2023cut, wang2024videocutler, huang2024uvis} methods have been proposed with different kinds of supervisions. Some methods rely on classification labels \cite{ahn2019weakly, liu2021weakly} and use the optical flow \cite{teed2020raft, eslami2024rethinking} methods for one-to-one correspondence. But, this kind of supervision is not sufficient for achieving higher performance. On the other hand, a variety of box-supervised methods have been proposed for image-level instance segmentation \cite{hsu2019weakly, lan2021discobox, tian2021boxinst, li2022box, li2023sim}. Although those methods achieve great performance due to the existence of large-scale datasets, like COCO \cite{lin2014microsoft}, they are not completely appropriate for VIS, as they ignore the temporal information in videos. As such, their performance is lower when applied to the video. In fact, with a 3D spatio-temporal volume, some temporal consistency constraints can be utilized to improve the performance further. Recently, the MaskFree VIS \cite{ke2023mask} work proposed the first VIS method with bounding box supervision by applying temporal consistency loss to match patch-oriented one-to-many correspondences to handle the occlusion and homogenous region scenarios. Doing this reduces the gap between fully supervised and weakly supervised VIS.

In this work, we extend the idea of \cite{ke2023mask} and define two more temporal consistency constraints. Inspired by deep spectral methods \cite{melas2022deep}, we leverage eigenvalues derived from the Laplacian matrix of a graph adjacency matrix as a robust representation of structural and spatial properties. To do this, first, a Temporal Eigenvalue Loss (TEL) is defined as clip-oriented and aims at minimizing the difference between the eigenvalues of instances in consecutive frames. This regularization term reduces the temporal discontinuities and improves the overall quality and reliability of the segmentation output for more consistent and visually pleasing results. This loss function integrates easily into existing VIS methods without requiring additional network parameters. Furthermore, a Quality Cluster Coefficient (QCC) is proposed for clip-level spatio-temporal consistency without relying on ground-truth labels.  The proposed QCC measures the quality and discrimination power of dense spatio-temporal areas, enabling dynamic adjustment of the impact of both patch-oriented and clip-oriented constraints. Qualitative results are provided in Fig. \ref{fig: pull_figure}. To summarize, the main contributions of this work are as follows:

\begin{itemize}
\item Proposing a wealy-supervised VIS method called Eigen-Cluster VIS, which only needs weak supervision to achieve competitive accuracy without requiring mask annotations. Notably, because all the introduced modules are only present during the training phase and within the loss function, this method does not add any trainable parameters to the network, ensuring that the inference speed remains unaffected compared to the baseline. 

\item Introducing clip-oriented views beside patch-oriented views to ensure spatio-temporal consistency with QCC and TEL. The QCC uses K-means clustering and the Davies-Bouldin score for unsupervised feature discrimination, while the TEL ensures temporal coherence by minimizing the MAE between eigenvalues of adjacent frames. Together, these functions enhance the overall performance of the proposed VIS method.

\item Analyzing the effectiveness of the proposed method on the YouTube-VIS 2019/2021 and OVIS benchmarks, demonstrating that Eigen-Cluster VIS narrows the performance gap between fully-supervised and weakly-supervised VIS methods. 
\end{itemize}

\section{RELATED WORK}

\label{sec: re_work}

\subsection{Video Instance Segmentation}
Based on their application, existing VIS methods can be divided into two distinct categories: offline and online methods.

\subsubsection{Online VIS} 
Online VIS methods \cite{bertasius2020classifying, lin2021video, yang2021crossover, li2022improving, li2023tcovis, ying2023ctvis, heo2023generalized, athar2023tarvis} have made significant progress due to the availability of robust detectors and image-level instance segmentation methods. Most of those works, including \cite{yang2019video} as the first VIS method, attempt to extend segmentation models \cite{he2017mask, ke2021deep} by adding a tracking head. Due to their online nature, they can be applied to real-time scenarios such as video surveillance and autonomous driving. 

\subsubsection{Offline VIS} 
Offline VIS methods \cite{cao2020sipmask, liu2021sg, li2021spatial, li2023tube, li2022video}, on the other hand, by having access to all frames, can take advantage of temporal information and post-processing, making them suitable for applications such as video editing and content analysis. The STEm-Seg \cite{athar2020stem}, as one of the first offline VIS methods, uses clustering to track instances by viewing the input video as a spatio-temporal 3D volume.

Aside from the offline/online nature of VIS methods, transformer-based methods have recently emerged and achieved impressive results in this task \cite{hwang2021video, thawakar2022video, yang2022temporally, ke2022mask}. The \cite{wang2021end} work was one of the first to employ transformers for VIS. Other methods like Seqformer \cite{wu2022seqformer} and Mask2Former (M2F) \cite{cheng2021mask2former} introduced the frame query decomposition and masked attention, respectively. The IDOL \cite{wu2022defense}, while effective, relied on post-processing for tracking. Recent advancements, such as VITA \cite{heo2022vita} and DVIS \cite{zhang2023dvis}, take a more sophisticated approach. The VITA leverages temporal interactions between the frame-level object queries, and DVIS adopts a modular design, separating VIS into distinct stages of segmentation, tracking, and refinement.

\subsection{Mask-Free Video Instance Segmentation}
Existing weakly-supervised methods for mask-free instance segmentation primarily operate at image level \cite{khoreva2017simple, lan2021discobox, tian2021boxinst, lee2021bbam, li2022box, li2023sim}. Early works, such as \cite{dai2015boxsup, hsu2019weakly}, relied on region proposals, leading to slow training times. More recently, point-supervised instance segmentation methods have emerged \cite{cheng2022pointly, lee2024extreme}. The BoxTeacher \cite{cheng2023boxteacher} work leverages a sophisticated teacher model to generate high-quality masks as pseudo-labels. While these methods demonstrate promising performance in weakly-supervised instance segmentation, they ignore temporal information presented in videos, as they operate solely at the image level.

Several works have been conducted in the areas of weakly-, semi-, and unsupervised VIS. Methods like flowIRN \cite{liu2021weakly} leverage classification labels alongside optical flow to enforce temporal mask consistency, while Cluster2Former \cite{fothi2024cluster2former} utilizes scribble-based annotations. The SOLO-Track \cite{fu2021learning} proposes a novel semi-supervised method for VIS that requires no video annotations, and MinVIS \cite{huang2022minvis} utilizes a small subset of annotated frames to train a query-based instance segmentation model with bipartite matching for mask tracking. In the unsupervised setting, VideoCutLER \cite{wang2024videocutler} employs MaskCut \cite{wang2023cut} to generate object pseudo-masks. It proposes ImageCut2Video to create synthetic videos using a pair of images and then trains an unsupervised VIS model using these mask trajectories. The most recent work, UVIS \cite{huang2024uvis}, uses DINO \cite{caron2021emerging} for mask generation and leverages CLIP \cite{radford2021learning} to add semantic information to these masks. It then trains a transformer to learn queries for predicting per-frame instance masks and introduces tracking banks to associate masks between frames.

The most relevant methods related to this work are Mask-Free VIS \cite{ke2023mask} and PM-VIS \cite{yang2024pm}. The Mask-Free VIS incorporates \cite{tian2021boxinst} as its instance segmentation baseline and introduces a novel temporal consistency constraint called Temporal KNN-patch Loss. This constraint is achieved by finding one-to-many patch matches across frames followed by k-nearest neighbor selection. A consistency loss is then applied to these matched patches. On the other hand, the PM-VIS employs different methods to generate three distinct pseudo-masks, which are filtered to achieve high-quality pseudo-masks. Finally, a fully-supervised VIS method is trained on these filtered pseudo-masks.

\subsection{Deep Spectral Methods}
The spectral graph theory, initially introduced in \cite{cheeger1970lower}, has evolved to connect global graph features by the eigenvalues and eigenvectors of the Laplacian matrix \cite{donath1973lower, fiedler1973algebraic}. Notably, \cite{donath1973lower} proposed the Fidler eigenvalue, the second smallest eigenvalue, as a measure of graph connectivity. Furthermore, \cite{fiedler1973algebraic} demonstrated the use of graph Laplacian eigenvectors for achieving energy-minimizing graph partitions. With the advent of machine learning and computer vision, \cite{shi2000normalized} and \cite{ng2001spectral} pioneered the application of spectral clustering for image segmentation. Inspired by these works, a recent method like DSM \cite{melas2022deep} and \cite{wang2022self}, leverage features extracted from a self-supervised transformer-based backbone, along with the Laplacian of the affinity matrix and patch-wise similarities, to perform tasks such as object localization and semantic segmentation. 

Building upon DSM \cite{melas2022deep}, research has primarily focused on object localization \cite{wang2022self, simeoni2021localizing, wang2023cut}, semantic segmentation \cite{li2023acseg, zadaianchuk2022unsupervised, aflalo2023deepcut}, and video object segmentation \cite{wang2023tokencut, ponimatkin2023simple, choudhury2022guess}. In contrast, instance segmentation has received less attention. The MaskCut \cite{wang2023cut} method constructs a patch-wise similarity matrix by using a self-supervised backbone. It then iteratively applies Normalized Cuts to generate foreground masks and masking out affinity matrix values for each discovered object. The work in \cite{arefi2024deep} has a similar paradigm to the DSM and shows that not all channels of a self-supervised backbone are useful for instance segmentation purposes. It proposes channel pruning methods for improving instance segmentation performance.

\section{PROPOSED METHOD}
\label{sec: prop_method}

\begin{figure*}[!ht]
	\centering
	\centerline{\includegraphics[width=\textwidth]{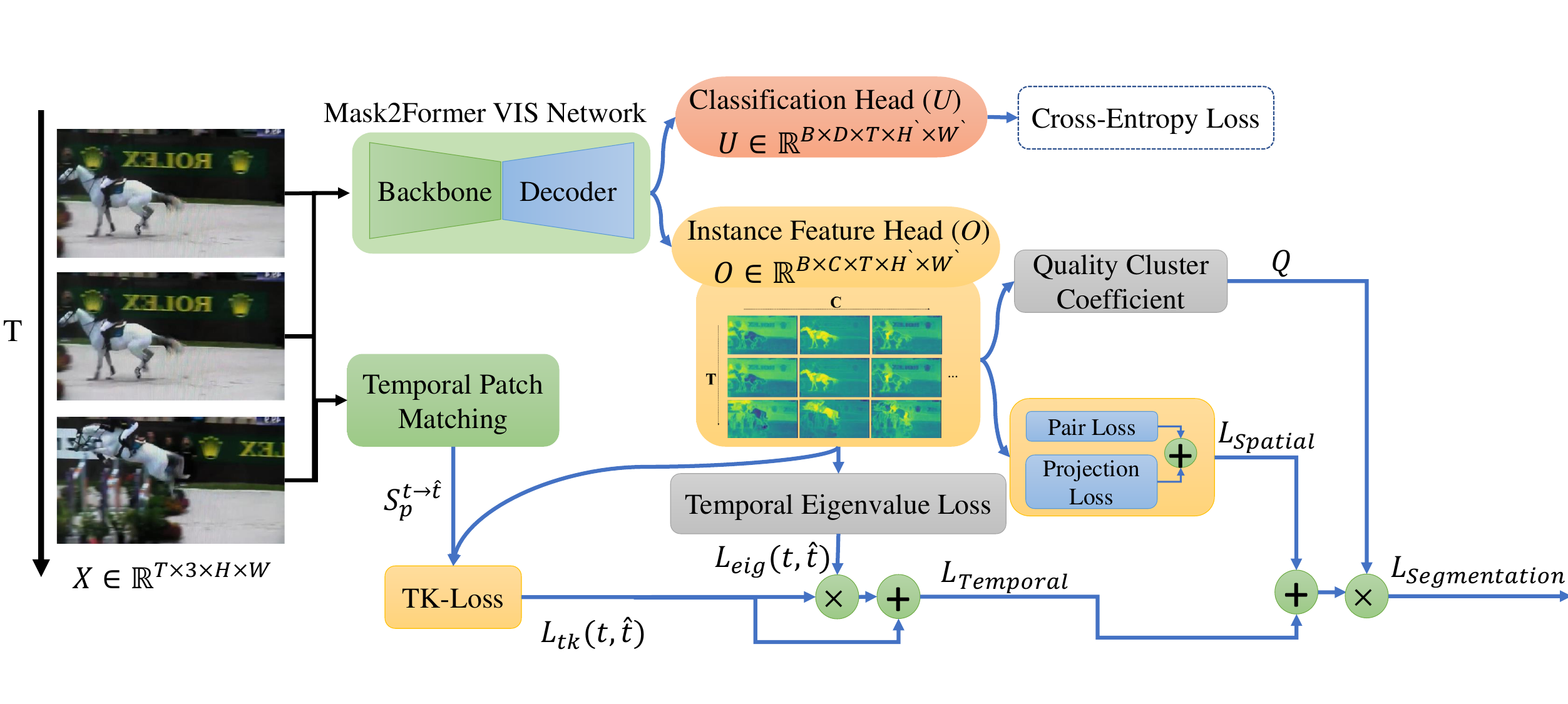}}
	\caption{{\bf Pipelines of proposed Eigen-Cluster VIS. After inputting \(X\) into Mask2former, two outputs, \(U\) and \(O\), are generated. According to Method \cite{ke2022mask}, \(X\) is also fed into the Temporal Patch Matching module to extract the corresponding patches for TK-Loss. \(U\) pertains to classification and is optimized using Cross-Entropy Loss, while \(O\) represents instance features from which the QCC, TEL, and TK-Loss modules are derived. All the losses are then aggregated under the title \(L_{\text{segmentation}}\) during the training process, and the optimization is carried out.}}
	\label{fig: Diagram}
\end{figure*}

Inspired by the weakly supervised approaches, this paper introduces an Eigen-Cluster VIS method, which performs segmentation without requiring mask annotation during the training phase. Eigen-Cluster VIS comprises two main components: QCC and TEL. Fig. \ref{fig: Diagram} shows their placement in the overall pipeline. The QCC measures the spatio-temporal coherence at the feature representation level in a video clip. Considering the eigenvalue components of predicted instances over time, the TEL predicts the masks with smoother transitions.
The proposed method will be briefly reviewed below, followed by a more detailed introduction of the specific modules.

\subsection{Overview}

Fig. \ref{fig: Diagram} illustrates the overall pipeline of the proposed method. It is worth mentioning that the proposed method can be integrated with any backbone and supervised VIS model. In this research, the M2F VIS is used with ResNet50-101 and Swin-L backbones. The input to the network is denoted as $X \in \mathbb{R}^{B \times T \times 3 \times H \times W}$, where $B$ represents the number of three-channel color clips, $T$ is the number of frames, and $H$ and $W$ are the height and width of each frame, respectively.

To apply the Temporal KNN-patch Loss (TK-Loss), as described in \cite{ke2023mask}, candidate patches must first be extracted from the input clip $X$. Then, the temporal matching is performed between these patches. If $p$ represents a patch location in frame $t$, the temporal matching yields the mapping $\mathcal{S}$, where $\mathcal{S}_p^{t \rightarrow \hat{t}} = \{ \hat{p}_i \}_i$ provides the corresponding locations $\hat{p}$ in frame $\hat{t}$.

After obtaining the corresponding patches, the $X$ clip is fed into the M2F network. The M2F produces two outputs: $O \in \mathbb{R}^{B \times C \times T \times H' \times W'}$, which contains the features corresponding to the instances, and $U \in \mathbb{R}^{B \times D \times T \times H \times W}$, which is the classification matrix corresponding to $D$ categories. The classification matrix $U$ is optimized by using the cross-entropy function and the ground truth targets.

During the training phase, the losses and coefficients related to predicted instances in space and time are calculated from the output $O$. These calculations occur at different levels: i) at the spatio-temporal level, which includes the QCC; at the temporal level, which includes the TEL and TK\_Loss; and ii) at the spatial level, which includes $L_{projection}$ and $L_{pair}$. The network weights are then optimized according to these criteria.

In the following sections, first the concept of spatio-temporal consistency is presented as it pertains to this research, followed by a detailed introduction of the proposed loss functions and coefficients for the VIS task.

\subsection{Spatio-Temporal Consistency}

In the past, spatial compatibility modules often used a patch-oriented approach to address compatibility issues. A patch-oriented view, such as the TK-Loss approach, focuses on small areas like corresponding patches across frames. This can be achieved by working with predicted masks \cite{lee2019frame, liu2021weakly} or using optical flow techniques as shown in \cite{tokmakov2016learning, tsai2016semantic}. However, in addition to the patch-oriented view, a clip-oriented view examines compatibility over larger areas.

In this research, the QCC module is introduced to create a clip-oriented view. This module measures spatio-temporal coherence at the feature representation level of a video clip without requiring any knowledge of the ground truth mask regions.
Although the patch-oriented view, like TK-Loss, strongly constrains the consistency of the mask for corresponding patches of instances across frames, it applies this constraint equally across all patches without accounting for inherent differences between frames. 

In successive frames, instances may be deformed or occluded, or their appearance may change due to motion blur and brightness. Therefore, the degree of inter-frame changes in each clip varies and can be measured and applied in the loss function at the level of patches. With the TEL module, we aim to create a regularizing coefficient between frames based on the difference in eigenvalues. This coefficient controls the penalty amount in the loss at the level of corresponding patches. As shown in Fig. \ref{fig: Diagram}, this controller increases the patch loss by multiplying the TK-Loss value that measures the loss at the patch level. Therefore, the more significant the difference in eigenvalues between frames, the greater the final temporal loss and, ultimately, the greater the effort to improve the accuracy of such frames. As demonstrated in Section 4, using eigenvalues as the controller coefficients effectively enhances the segmentation accuracy.

\subsection{Quality Cluster Coefficient}

During the training process, specifically in the loss calculation stage, the output $O$ converges towards a suitable representation of each instance based on the ground truth mask. However, beyond just training with ground truth masks, it is possible to design a loss function that measures the discrimination of the features obtained without relying on the ground truth. This allows for an unsupervised evaluation of the output quality of $O$.
Incorporating an unsupervised perspective enables the model to learn new aspects that might not be captured by the supervised view alone. This approach allows for better adaptation when the distribution of objects during testing differs significantly from that during training. By learning aspects independently of the ground truth and focusing more on the internal consistency of the output, the model can improve mask prediction accuracy and other related tasks.

\begin{figure*}[!ht]
	\centering
	\centerline{\includegraphics[width=\textwidth]{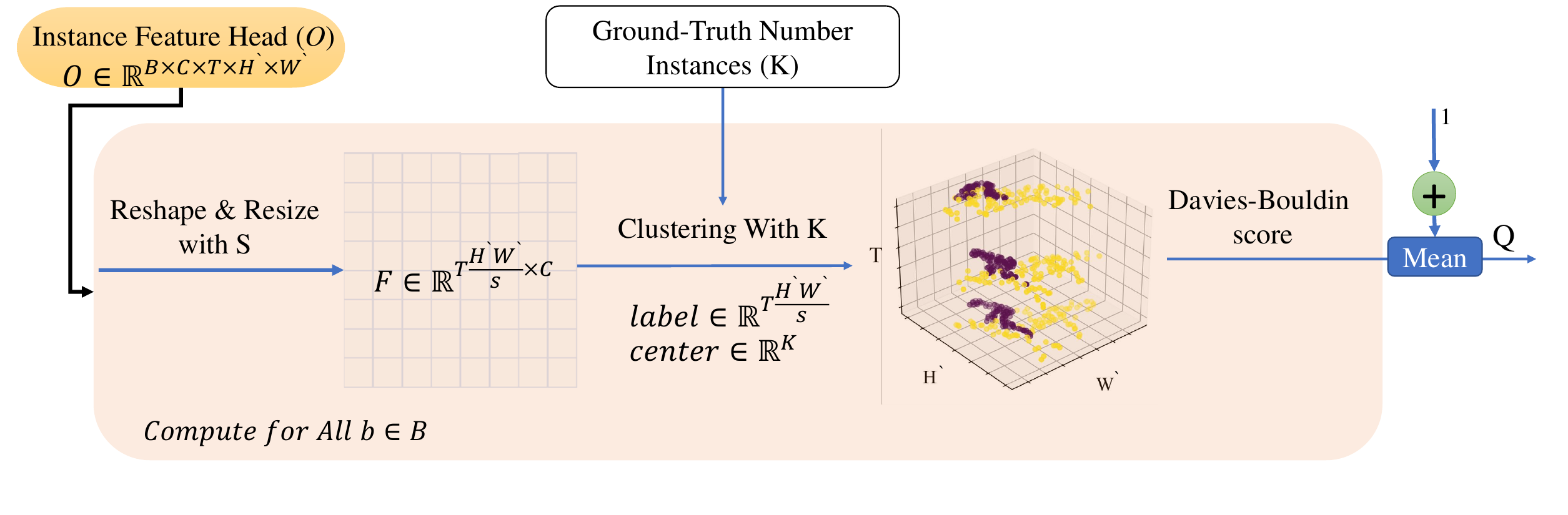}}
	\caption{{\bf Quality Cluster Coefficient. Instance features are reshaped and resized along the spatial dimension for each clip in a batch. Subsequently, a K-means clustering (where K is the number of instances in the ground truth) is performed on these features. The Quality Cluster Coefficient is computed by using the Davies-Bouldin metric.}}
	\label{fig: QCC}
\end{figure*}

To achieve this, a new metric called the QCC is introduced, which emphasizes the natural grouping consistency of features and promotes better segmentation of instances. The QCC as shown in Fig. \ref{fig: QCC} is calculated by applying the K-means clustering (with K equal to the number of real instances) on $O$ and evaluating the quality of the resulting spatio-temporal clusters by using the Davies-Bouldin score \cite{davies1979cluster}.
By implementing this coefficient, the mismatch between the number of clusters and the type of clustering is reduced, allowing the model to converge towards denser clusters with a higher resolution gradually. This leads to better discrimination of the features representing the instances. The QCC measures the quality and discrimination of dense areas in space and time without considering the actual regions of the instances. This nearly  unsupervised approach allows the network to self-evaluate its outputs, which can uncover new learning opportunities when combined with other loss functions that provide feedback from the real output. Note that this score is straightforward to implement and does not add any new parameters to the network. Further details on the main steps are explained in the follow:

\textbf{1. Feature Transformation:} Each instance feature \(O \in \mathbb{R}^{B \times C \times T \times H' \times W'}\) is first resized using the scale factor \(S\). Then, each batch is resized to the dimensions \(F \in \mathbb{R}^{(T H' W' / S) \times C}\).

\textbf{2. K-means Clustering:} For each batch, the K-means algorithm is applied to cluster the transformed features into \(K\) clusters. The K-means algorithm partitions \(F\) into \(K\) clusters \(\{V_1, V_2, \ldots, V_K\}\) by minimizing the within-cluster variance. In this context, \(f\) denotes an individual feature vector in \(F\). The objective is defined as:
\begin{align}
	\label{Eq1: clustering formula}
	\min_{\{V_k\}} \sum_{k=1}^K \sum_{f \in V_k} \|f - \mu_k\|^2
\end{align}

\noindent where \(\mu_k\) is the centroid of cluster \(V_k\).

\textbf{3. Davies-Bouldin Score:} The quality of the clustering is evaluated using the Davies-Bouldin score, which is defined as:
\begin{align}
	\label{Eq2: davis-bouldin formula}
	d_b = \frac{1}{K} \sum_{i=1}^K \max_{j \ne i} \left( \frac{\sigma_i +  \sigma_j}{dist(\mu_i, \mu_j)} \right) , b \in B
\end{align}

\noindent where \(\sigma_i\) is the average distance between each point in cluster \(i\) and the centroid \(\mu_i\). Also, \(dist(\mu_i, \mu_j)\) is the Euclidean distance between centroids \(\mu_i\) and \(\mu_j\). Lower \(d_b\) scores indicate better cluster quality.

\textbf{4. Creating the Coefficient:} Since QCC is considered a coefficient, the sum of the \(d_b\) values plus one is chosen as the final value of \(Q\):
\begin{align}
	\label{Eq3: QCC metric}
	Q = 1 +  \frac{1}{B} \sum_{b \in B} d_b,
\end{align}

\noindent which is then multiplied by the final loss. A lower value of \(Q\) encourages the network to learn better features, leading to more distinct clusters.

\subsection{Temporal Eigenvalue Loss}

\begin{figure*}[!ht]
	\centering
	\centerline{\includegraphics[scale=0.35]{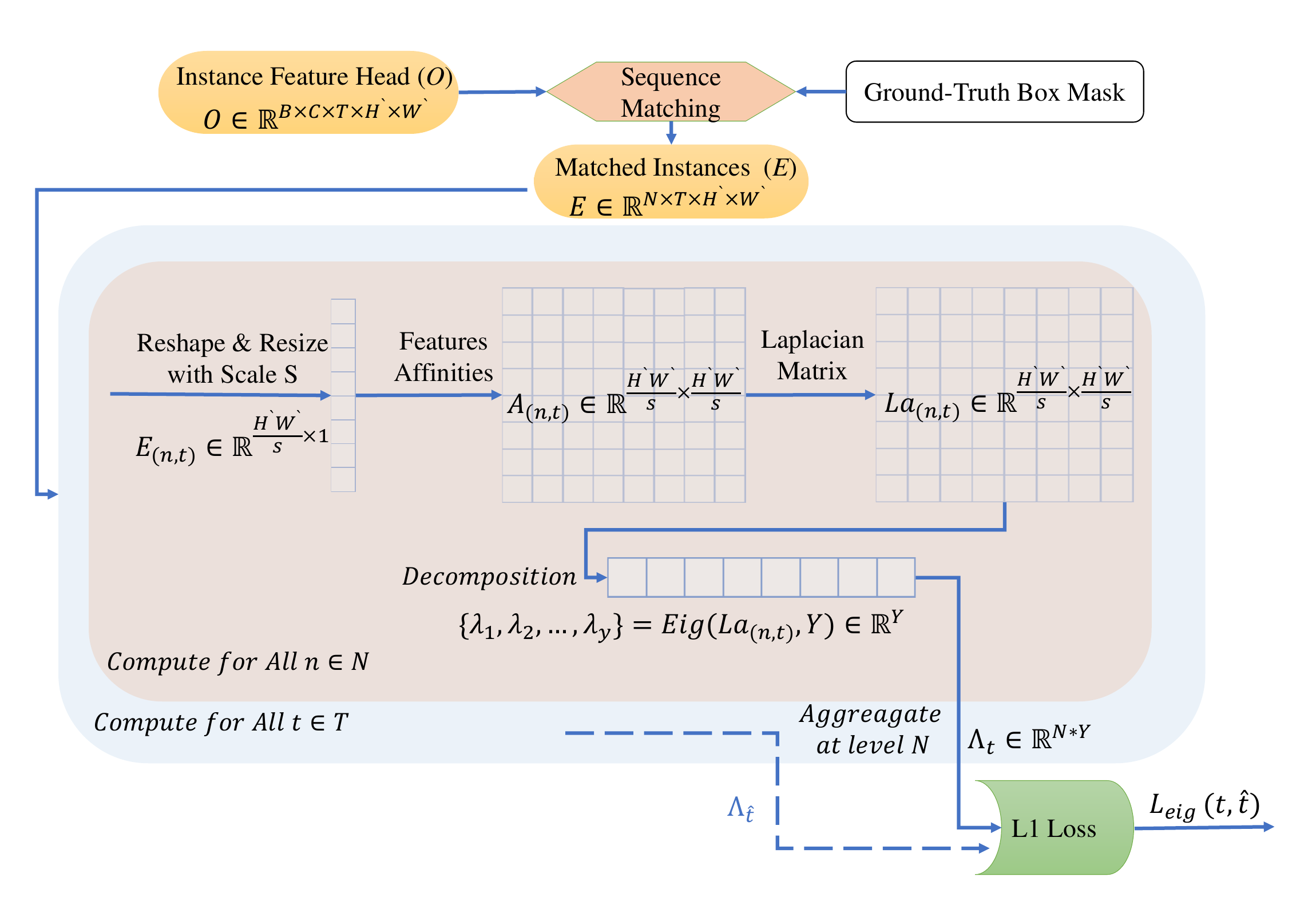}}
	\caption{{\bf Temporal Eigenvalue Loss. Matched instances are identified through sequence matching between the instance features and ground truth masks. Subsequently, for each instance in every frame, the features are resized and reshaped, followed by the creation of an affinity matrix based on these features. Eigenvalues of the Laplacian of this affinity matrix are extracted for each instance. The MAE loss is then computed cyclically between the corresponding instances in every two consecutive frames.}}
	\label{fig: TEL}
\end{figure*}

The goal of TEL, shown in Fig. \ref{fig: TEL}, is to ensure temporal consistency and coherence in VIS. As explained earlier, the eigenvalues derived from the Laplacian matrix of a graph adjacency matrix provide a robust representation of the structural and spatial properties in each graph. These eigenvalues offer essential information about the connectivity and smoothness of pixel links in each frame, making them useful in the graph segmentation process.
For example, the second smallest eigenvalue, known as the algebraic connectivity or Fiedler value, reflects the algebraic connectivity and is related to the sparsest cut and spectral gap. Other eigenvalues can represent the structure of the graph at higher levels of abstraction.

By converting the output $O$ to an adjacency matrix, the characteristics of eigenvalues can be utilized in the context of input video frames.
Ensuring that eigenvalues are close to each other between successive frames is crucial for maintaining temporal coherence. Time warping in segmentation tasks helps to avoid flickering and non-contiguous segmentation in frames, which are common problems in video analysis. Minimizing the MAE between eigenvalues of adjacent frames promotes smoother transitions and more stable segmentation boundaries over time.

This alignment of eigenvalues across frames reduces temporal discontinuities and increases the overall quality and reliability of the segmentation output, resulting in more consistent and visually pleasing results. One of the important features of this loss function is the small dimensions of the compared data, making the optimization process easier and not imposing an additional burden on the overall network optimization process. Note that this loss function does not add any parameters to the network. The main steps of this loss function are outlined below.

\textbf{1. Select the Corresponding Pair:} For each real instance $i \in \{1, \ldots, I\}$ in batch $b \in \{1, \ldots, B\}$, we determine the corresponding channel $c_i$ from the prediction matrix $O$ by using the Hungarian algorithm \cite{kuhn1955hungarian}. This result is a set of pairs $R = \{(i, c_i)\}_{i=1}^I$, where $c_i$ is the selected channel for instance $i$. The total number of pairs $(i, c_i)$ across all batches is $N = B \times I$. Using these pairs $R$, a target matrix $E \in \mathbb{R}^{N \times T \times H' \times W'}$ is constructed. The elements of $E$ are obtained by indexing $O$ with the selected channels from $R$.

\textbf{2. Feature Transformation and Affinity Matrix Construction:} For each $t \in T$ and each $n \in N$, we first resize the $E$ matrix by a scale factor $S$ and then reshape it as follows:
\begin{align}
	\label{Eq4: scale_reshape}
	E_{(n,t)} \in \mathbb{R}^{\frac{H'W'}{S} \times 1}.
\end{align}

Next, the affinity matrix $A_{(n,t)}\in \mathbb{R}^{\frac{H'W'}{S} \times \frac{H'W'}{S}}$ is constructed by:
\begin{align}
	\label{Eq5: affinity matrix construction}
	A_{(n,t)} = E_{(n,t)} E_{(n,t)}^T \odot \left(E_{(n,t)} E_{(n,t)}^T > 0 \right)
\end{align}

\noindent where $\odot$ denotes the element-wise multiplication. The affinities are thresholded at $0$ because the features are intended for aggregating similar features rather than anti-correlated features.

\textbf{3. Building the Laplacian Matrix and Decomposing it into Eigenvalues:} First, the Degree Matrix $D_{(n,t)}$ is calculated as:
\begin{align}
	\label{Eq6: matrix degree}
	D_{(n,t)} = \sum_j A_{(n,t)}(i,j).
\end{align}

Then, the Laplacian matrix $La_{(n,t)}$ is constructed as:
\begin{align}
	\label{Eq7: laplacian matrix}
	La_{(n,t)} = D_{(n,t)} - A_{(n,t)}.
\end{align}

Finally, \(La_{(n,t)}\) is decomposed to obtain the first \(Y\) eigenvalues, resulting in:
\begin{align}
	\label{Eq8: eigen values}
	\{\lambda_1, \lambda_2, \ldots, \lambda_Y\} = \text{Eig}(La_{(n,t)},Y) \in \mathbb{R}^Y.
\end{align}

\textbf{4. Calculation of Loss:} All eigenvalues are stacked at the level of each $n \in N$ to get $\Lambda_t$:
\begin{align}
	\label{Eq9: eigen values stack}
	\Lambda_t = \text{Stack}([\lambda_1, \lambda_2, \ldots, \lambda_Y]) \in \mathbb{R}^{N \times Y}.
\end{align}

Then, the eigenvalue loss $\mathcal{L}_{\text{eig}}(\hat{t},t)$ is calculated using the following relationship:
\begin{align}
	\label{Eq9: eigen values loss}
	\mathcal{L}_{\text{eig}}(t, \hat{t}) = \| \Lambda_{\hat{t}} - \Lambda_{t} \|_1.
\end{align}

\subsection{Segmentation Loss}

As explained in \cite{ke2022mask}, the TK-loss, denoted as \( L_{\text{tk}}(t, \hat{t}) \), is derived by considering the matrix \( O \), the locations \( p \), and the mapping \( \mathcal{S}_p^{t \rightarrow \hat{t}} \). This can be expressed more simply as:
\begin{align}
	\label{Eq10: TK-Loss}
	\tau = M^t_p M^{\hat{t}}_{\hat{p}} +  (1 - M^t_p)(1 - M^{\hat{t}}_{\hat{p}}),
\end{align}

\begin{align}
	\label{Eq11: TK-Loss 2}
	L_{\text{tk}}(t, \hat{t}) = \frac{1}{H'W'} \sum_{p} \sum_{\hat{p} \in \mathcal{S}^{t \rightarrow \hat{t}}} -\log(\tau),
\end{align}

\noindent where \( M^t_p \in [0,1] \) represents the predicted binary instance mask of an object, evaluated at position \( p \) in frame \( t \).

According to the given explanations, \( L_{\text{eig}}(t, \hat{t}) \) is first multiplied by \( L_{\text{tk}}(t, \hat{t}) \), and then added again with \( L_{\text{tk}}(t, \hat{t}) \) to finally obtain the temporal loss \( L_{\text{temporal}} \), as:
\begin{align}
	\label{Eq12: temporal loss}
	\psi(t, \hat{t}) = L_{\text{tk}}(t, \hat{t})+ \beta \left(L_{\text{tk}}(t, \hat{t}) \times L_{\text{eig}}(t, \hat{t})\right),
\end{align}

\begin{align}
	\label{Eq13: temporal loss 2}
	L_{\text{temporal}} = \sum_{t=1}^{T} 
\begin{cases} 
\psi(t, t+1) & \text{if } t < T-1 \\ 
\psi(t+1, 0) & \text{if } t = T-1 
\end{cases},
\end{align}

\noindent where $\beta$ is a hyperparameter chosen according to the complexity of the backbone.

Since the proposed method is based on weak supervision and does not involve ground truth masks, we draw inspiration from \cite{tian2021boxinst} to estimate a spatial loss, \( L_{\text{spatial}} \), using two components: the Box Projection Loss \( L_{\text{proj}} \) and the Pairwise Loss \( L_{\text{pair}} \). The Box Projection Loss \( L_{\text{proj}} \) measures the compatibility between the predicted and ground truth bounding boxes by projecting the predicted and ground truth masks along the $x$ and $y$ axes, using the Dice loss function. The concept behind \( L_{\text{pair}} \) is that pixels with similar colors most probably belong to the same predicted mask.

After obtaining the spatial loss and temporal loss and multiplying them by \( Q \), which indicates the clustering quality in space-time, the final segmentation loss \( L_{\text{seg}} \) can be computed as:

\begin{align}
    L_{\text{segmentation}} &= Q^{\alpha} \times \left( L_{\text{spatial}} + L_{\text{temporal}} \right)
    \label{Eq15:segmentation-Loss}
\end{align}

\noindent where $\alpha$ is a value of zero or one used in the training process, as described in Section 4.

As described in \cite{ke2023mask}, during the instance-sequence matching step, the predicted and ground truth masks are first converted into box masks. Then, random points are sampled from these masks using the Dice IoU loss function to identify the best pairs of predicted and ground truth masks.

\section{EXPERIMENTS}
\label{sec: experiments}

This section begins by introducing the experimental datasets and evaluation metrics. It then explains the implementation methods. Finally, the experiments and ablation studies are presented and discussed.

\begin{table*}[ht]
	\caption{\label{tab:vis2019} {\bf Comparison on the YouTube-VIS 2019 validation dataset: \textbf{I} refers to the use of the COCO dataset during pre-training at the image mask level, while \textbf{V} indicates that video masks were used during the training process.\\}}
	\centering
    
	\begin{tabular}{l c c c c c c c c}
		\toprule
		Method & Mask ann.  & Backbone & AP & $AP_{50}$ & $AP_{75}$ & $AR_1$ & $AR_{10}$ \\
		
		\midrule
		
		  \textit{Fully-supervised:} & & & & & & & \\ 
            
            PCAN \cite{ke2021prototypical} & I+V & Res50 & 36.1 & 54.9 & 39.4 & 36.3 & 41.6 \\
            EfficientVIS \cite{wu2022efficient} & I+V & Res50 & 37.9 & 59.7 & 43.0 & 40.3 & 46.6 \\
            InsPro \cite{he2022inspro} & I+V & Res50 & 40.0 & 62.9 & 43.1 & 37.6 & 44.5 \\
            IFC \cite{hwang2021video} & I+V & Res50 & 42.8 & 65.8 & 46.8 & 43.8 & 51.2 \\
            SeqFormer \cite{wu2022seqformer} & I+V & Res50 & 47.4 & 69.8 & 51.8 & 45.5 & 54.8  \\
            Mask2Former \cite{cheng2021mask2former} & I+V & Res50 & 47.8 & 69.2 & 52.7 & 46.2 & 56.6 \\
            VMT \cite{ke2022video} & I+V & Res50 & 47.9 & - & 52.0 & 45.8 & - \\
            GenVIS \cite{heo2023generalized}& I+V & Res50 & 50.0 & 71.5 & 54.6 & 49.5 & 59.7 \\
            DVIS \cite{zhang2023dvis} & I+V & Res50 & 51.2 & 73.8 & 57.1 & 47.2 & 59.3 \\ 

		\midrule
		
		\textit{Weakly-supervised:} & & & & & & & \\ 
            FlowIRN \cite{liu2021weakly} & - & Res50 & 10.5 & 27.2 & 6.2 & 12.3 & 13.6 \\
            SOLO-Track \cite{fu2021learning} & I & Res50 & 30.6 & 50.7 & 33.5 & 31.6 & 37.1 \\
            Cluster2Former \cite{fothi2024cluster2former} & I & Res50 & 38.3 & 62.5 & 42.5 & 39.3 & 46.6\\
            Mask2Former + BoxInst \cite{tian2021boxinst} & - & Res50 & 38.6 & 64.2 & 38.5 & 38.0 & 46.8 \\
            Mask2Former + Mask-free VIS \cite{ke2023mask} & I & Res50 & 46.6 & \textbf{72.5} & 49.7 & 44.9 & 55.7 \\
            Mask2Former + \textbf{Eigen-Cluster VIS} & I & Res50 & \textbf{47.5} & 72.1 & \textbf{51.1} & \textbf{45.7} & \textbf{56.3} \\

            \midrule \midrule

            \textit{Fully-supervised:} & & & & & & & \\ 
            PCAN \cite{ke2021prototypical} & I+V & Res101   & 37.6 & 57.2 & 41.3 & 37.2 & 43.9 \\
            EfficientVIS \cite{wu2022efficient} & I+V & Res101   & 39.8 & 61.8 & 44.7 & 42.1 & 49.8 \\
            IFC \cite{hwang2021video} & I+V & Res101   & 44.6 & 69.2 & 49.5 & 44.0 & 52.1 \\
            SeqFormer \cite{wu2022seqformer} & I+V & Res101 & 49.0 & 71.1 & 55.7 & 46.8 & 56.9  \\
            VMAT  \cite{ke2022video} & I+V & Res101 & 49.4 & - & 56.4 & 46.7 & - \\
            Mask2Former \cite{cheng2021mask2former} & I+V & Res101 & 49.8 & 73.6 & 55.4 & 48.0 & 58.0  \\

            \midrule
  
		\textit{Weakly-supervised:} & & & & & & & \\ 
            Mask2Former + BoxInst \cite{tian2021boxinst} & - & Res101 & 40.8 & 67.8 & 42.2 & 40.0 & 49.2 \\
            Mask2Former + Mask-free VIS \cite{ke2023mask} & I & Res101 & 48.9 & 74.9 & \textbf{54.7} & 44.9 & 57.0 \\
            Mask2Former + \textbf{Eigen-Cluster VIS} & I & Res101 & \textbf{49.8} & \textbf{76.1} & 54.2 & \textbf{46.8} & \textbf{58.4} \\

            \midrule \midrule

            \textit{Fully-supervised:} & & & & & & & \\ 
            SeqFormer \cite{wu2022seqformer} & I+V & SwinL & 59.3 & 82.1 & 66.4 & 51.7 & 64.4 \\
            VMAT \cite{ke2022video} & I+V & SwinL &  59.7 & - & 66.7 & 52.0 & - \\
            Mask2Former \cite{cheng2021mask2former} & I+V & SwinL & 60.4 & 84.4 & 67.0 & - & - \\
            DVIS \cite{zhang2023dvis} & I+V & SwinL & 63.9 & 78.2 & 70.4 & 56.2 & 69.0 \\
            GenVIS \cite{heo2023generalized} & I+V & SwinL & 64.0 & 84.9 & 68.3 & 56.1 & 69.4 \\ 

            \midrule
		
		\textit{Weakly-supervised:} & & & & & & & \\ 
            Mask2Former + BoxInst \cite{tian2021boxinst} & - & SwinL & 49.8 & 73.2 & 55.5 & 48.2 & 58.1 \\
            Mask2Former + Mask-free VIS \cite{ke2023mask} & I & SwinL & 55.3 & 82.5 & 60.8 & 50.7 & 62.2 \\
            Mask2Former + \textbf{Eigen-Cluster VIS} & I & SwinL & \textbf{57.3} & \textbf{84.5} & \textbf{62.8} & \textbf{51.9} & \textbf{64.3} \\

		\bottomrule 
	\end{tabular}
\end{table*}

\subsection{Dataset and Evaluation Metrics}

\subsubsection{Datasets} 
Two datasets, YouTube-VIS 2019/2021 \cite{yang2019video} and OVIS \cite{qi2022occluded}, were used for experiments in the VIS task. YouTube-VIS 2019 contains 2,238 videos across 40 categories, mainly featuring animals, humans, and sports activities. In the 2021 version, the dataset has expanded to 3,032 videos, with some class categories differing from the previous version. OVIS, a VIS dataset with 607 videos across 25 categories, focuses on animals and humans. It presents more challenging conditions than the YouTube-VIS datasets due to a higher number of instances per video and greater occlusion, measured by MBOR \cite{qi2022occluded}. The instance-to-video ratio is 1.69 for YouTube-VIS 2019, 2.10 for YouTube-VIS 2021, and 5.80 for OVIS. The MBOR values are 0.07 and 0.06 for YouTube-VIS 2019 and 2021, respectively, and 0.22 for OVIS. 

\subsubsection{Evaluation Metrics} 
Following the COCO \cite{lin2014microsoft} guidelines and methodology of \cite{yang2019video}, the evaluation metrics are based on AP, AP50, AP75, AR1, and AR10.

\subsection{Implementation Details}

As mentioned earlier, the M2F network is used as the baseline for the VIS task in this research. All settings of this network are preserved, except for the loss function and sequence matching components. The loss function follows the TK-Loss settings as outlined in \cite{ke2023mask}. The training process is divided into three stages. The first stage involves pre-training with the COCO dataset using image-level masks. In contrast, the second and third stages involve the main training process, where no ground-truth masks are available, and only bounding box information is used. 

In the first and second stages, all backbone and network head weights are optimized with \( \alpha = 0 \) and \( \beta = 0 \) settings. In the third stage, the backbone is frozen, and \( \beta \) is adjusted based on backbone complexity settings, with \( \alpha \) set to 1. The batch size is 16 in the first and second stages and 2 in the third stage. The number of iterations is 60k for the first stage, 6k for the second stage, and up to 3.4k for the third stage, depending on different backbones. The AdamW optimization technique is used with a learning rate of \(10^{-4} \) in all three stages. In the QCC and TEL modules, we set \( S = 4 \) to reduce dimensions, and the number of extracted eigenvalues was \( Y = 3 \).

\subsection{Comparison with State-of-the-art Methods}

The proposed Eigen-Cluster VIS is compared to state-of-the-art methods in weak and fully supervised settings. Table \ref{tab:vis2019} presents the evaluation results on the YouTube-VIS 2019 validation set across three backbones: R50, R101, and SwinL.

For the R50 backbone, Eigen-Cluster VIS achieves an AP of 47.5, demonstrating clear superiority over fully supervised methods such as EfficientVIS\cite{wu2022efficient}, PCAN\cite{ke2021prototypical}, INSpro \cite{he2022inspro}, IFC \cite{hwang2021video}, and SeqFormer\cite{wu2022seqformer}. Additionally, in this backbone shows an improvement of 0.9\% over the Mask-freeVIS method and reduces the performance gap between Weakly-supervised methods and M2F \cite{cheng2021mask2former}  from 1.2 to 0.3.

With the R101 backbone, our proposed method achieves an AP of 49.8, outperforming fully supervised methods\cite{wu2022efficient},\cite{hwang2021video} and \cite{wu2022seqformer}. It also improves by 0.9\% over the MaskFreeVIS method, narrowing the gap between Weakly-supervised methods and M2F from 0.9 to 0.

In the SwinL backbone, which employs a transformer architecture, fully supervised methods generally exhibit higher accuracy than other backbones. However, our proposed method reduces the performance gap between Weakly-supervised methods and M2F from 5.1 to 3.1, achieving an AP of 57.3.

The designed modules cause only a 1.2x decrease in training speed compared to M2F \cite{cheng2021mask2former} (approximately two additional hours of training). However, there is no reduction in speed during the inference phase. Overall, the proposed method achieves segmentation speeds of 29.7 fps on R50, 29.2 fps on R101, and 14.9 fps on SwinL. These results were obtained using an NVIDIA 3090TI GPU.

\subsection{Ablation study}

The method presented in \cite{ke2022mask} provides a valuable comparison of various temporal matching schemes designed to enhance the performance of temporal matching algorithms in the VIS task, as shown in Table \ref{tab:temporal_matching}. These temporal matching schemes can be integrated into the loss function to ensure consistency between instance masks over time. As shown in Table \ref{tab:temporal_matching}, the proposed temporal matching achieved an AP of 44.2, improving the previous best matching method by 1.7\%.

A significant advantage of the matching algorithm presented in this research is that it introduces no additional learnable parameters. Furthermore, its clip-oriented design enables improved performance in occluded areas. In contrast, flow-based \cite{teed2020raft} algorithms struggle when an object’s pixels are occluded in other frames, leading to inaccurate predictions of pixel flow and consistency in those areas. Other matching algorithms, such as Deformable\cite{liu2021weakly} Matching and the set of Learnable Pixel/Patch Matching, 3D Pairwise Matching, and Temporal KNN-Matching introduced by \cite{ke2022mask}, provide only a patch-oriented perspective. In contrast, our proposed method also incorporates a clip-oriented view, resulting in higher accuracy than other methods.

\begin{figure}[!ht]
        \raggedleft
	\centering
	\centerline{\includegraphics[scale=0.25]{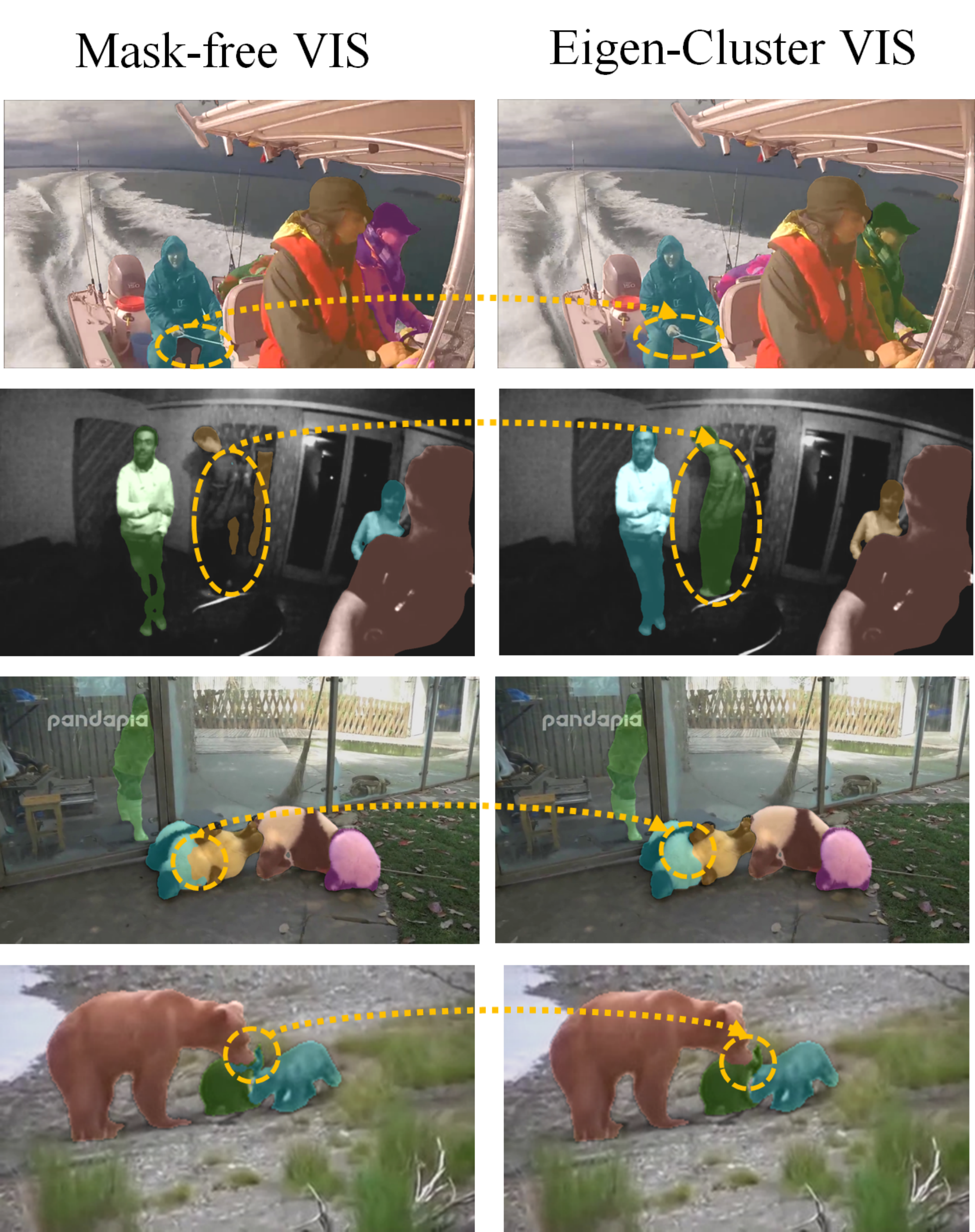}}
	\caption{{\bf Visualization of Baseline and Eigen-Cluster VIS Differences. In the baseline, some foreground areas may be missed, particularly in details involving fast movements. These missed regions are indicated with dashed lines.}}
	\label{fig:ablation}
\end{figure}

The effect of different modules, both individually and in combination, is shown in Table \ref{tab:metrics_tel_qcc_flipped}. Adding the QCC and TEL modules increases the AP by 0.8\% and 1.4\%, respectively. When both AP modules are added, the accuracy reaches 44.2, demonstrating the effectiveness of these modules. Fig. \ref{fig:ablation} illustrates the impact of adding and removing the proposed modules.

\begin{figure}[!ht]
        \raggedleft
	\centering
	\centerline{\includegraphics[scale=0.20]{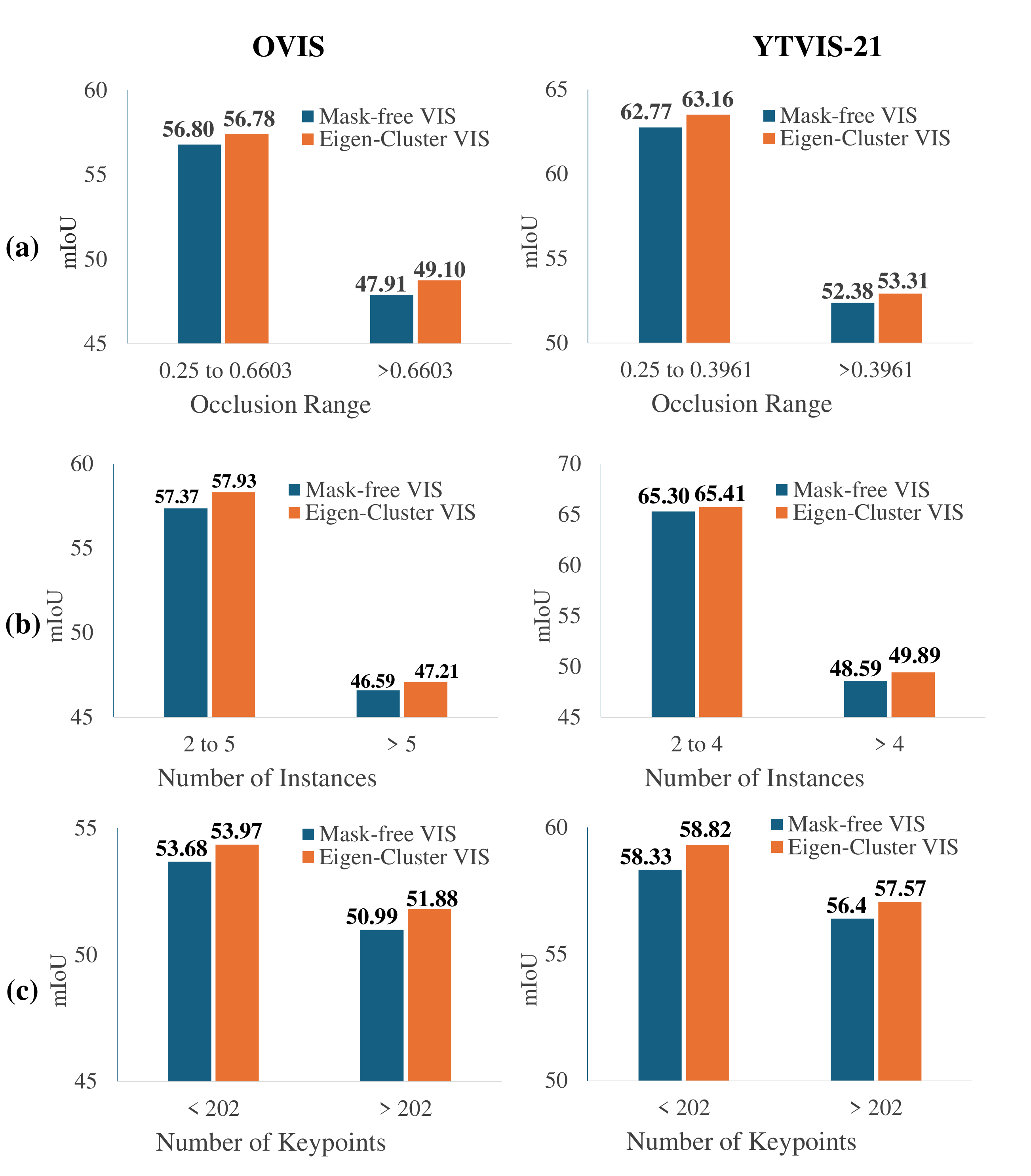}}
	\caption{{\bf mIoU evaluation across the YouTube-VIS 2021 (YTVIS-21) and OVIS datasets, examining three characteristics.}  (a) mIoU concerning occlusion, (b) mIoU in relation to the number of instances, and (c) mIoU based on the number of keypoints.}
	\label{fig:ablation23}
\end{figure}

In addition, Table \ref{table:impactY} shows the impact of $Y$ on segmentation accuracy. As evident, the best performance was achieved when ${Y = 3}$.

In Tables \ref{tab:temporal_matching}, \ref{tab:metrics_tel_qcc_flipped}, and \ref{table:impactY}, the R50 backbone is utilized, and only bounding box masks are employed at all stages of training. In contrast, our proposed results in Table \ref{tab:vis2019} leverage mask annotations during the first stage (image-level) of training.

\begin{table}[h!]
\centering
\caption{\bf{Comparison of Temporal Matching Schemes on Youtube-VIS 2019 validation set. The term \textbf{"Param"} refers to whether a learnable parameter is added to the model.}}
\resizebox{\linewidth}{!}{
\begin{tabular}{l@{\hskip 6pt}c@{\hskip 6pt}c@{\hskip 6pt}c@{\hskip 6pt}c@{\hskip 6pt}c@{\hskip 6pt}c}
\toprule
\textbf{Temporal Matching Scheme} & \textbf{Param} & \textbf{AP} & \textbf{AP50} & \textbf{AP75} & \textbf{AR1} & \textbf{AR10} \\
\midrule
Baseline (No Matching)            & \ding{55} & 38.6 & 65.9 & 38.8 & 38.4 & 47.7 \\
\midrule
3D Pairwise Matching              & \ding{55} & 39.4 & 65.0 & 41.7 & 40.2 & 48.0 \\
Learnable Pixel Matching          & \ding{51} & 39.5 & 65.7 & 40.2 & 39.7 & 48.4 \\
Temporal Deformable Matching      & \ding{51} & 39.6 & 65.9 & 40.1 & 39.9 & 48.6 \\
Flow-based Matching               & \ding{51} & 40.2 & 66.3 & 41.9 & 40.5 & 49.1 \\
Learnable Patch Matching          & \ding{51} & 40.6 & 66.5 & 42.6 & 40.3 & 49.2 \\
Temporal KNN-Matching (TK)        & \ding{55} & 42.5 & 66.8 & 45.7 & 41.2 & 51.2 \\
\midrule
\bfseries Temporal EigenValue + TK & \bfseries \ding{55} & \bfseries 44.2 & \bfseries 69.2 & \bfseries 47.2 & \bfseries 41.6 & \bfseries 52.0 \\
\bottomrule
\end{tabular}
}
\label{tab:temporal_matching}
\end{table}

\begin{table}[h!]
\centering
\caption{\bf{An ablation study to analyze the impact of proposed components on Youtube-VIS 2019 validation set.}}
\setlength{\tabcolsep}{8pt} 
\begin{tabular}{ccccccc}
\toprule
\textbf{TEL} & \textbf{QCC} & \textbf{AP} & \textbf{AP50} & \textbf{AP75} & \textbf{AR1} & \textbf{AR10} \\
\midrule
 & & 42.5 & 66.8 & 45.7 & 41.2 & 51.2 \\
 & \ding{51} & 43.3 & 68.3 & 45.9 & 41.1 & 51.0 \\
\ding{51} & & 43.9 & 70.5 & 47.2 & 41.6 & 51.7 \\
\ding{51} & \ding{51} & \textbf{44.2} & \textbf{69.2} & \textbf{47.2} & \textbf{41.6} & \textbf{52.0} \\
\bottomrule
\end{tabular}
\label{tab:metrics_tel_qcc_flipped}
\end{table}

\begin{table}[h!]
\centering
\caption{\bf{Impact of the number of eigenvalues Y on segmentation accuracy on Youtube-VIS 2019 validation set.}}
\setlength{\tabcolsep}{8pt} 
\begin{tabular}{cccccc}
\toprule
\textbf{Y} & \textbf{AP} & \textbf{AP50} & \textbf{AP75} & \textbf{AR1} & \textbf{AR10} \\
\midrule
2 & 43.7 & 68.9 & 46.6 & 41.5 & 51.5 \\
3 & \textbf{44.2} & \textbf{69.2} & \textbf{47.2} & \textbf{41.6} & \textbf{52.0} \\
4 & 43.4 & 67.8 & 46.6 & 41.3 & 51.5 \\
\bottomrule
\end{tabular}
\label{table:impactY}
\end{table}

Previous research has not reported the effect of the inherent characteristics of videos on the obtained accuracy, primarily because ground-truth validation tags have not been publicly released. As a result, it has been difficult to measure the accuracy of different methods based on various factors, such as occlusions, the number of instances, and the number of image keypoints.
To address this, we conducted an experiment to evaluate the accuracy of a model trained on YouTube-VIS2019 across different video attributes. The model was tested without exposure to the YouTube-VIS2021 and OVIS datasets during training. Since the classes used for training and testing are different, we only utilized the mIoU metric. Table \ref{table:otherdataset} presents the experimental results of the model trained on YouTube-VIS2019 and tested on YouTube-VIS2021 and OVIS datasets using the SwinL backbone. Due to the overlap of some videos between YouTube-VIS2021 and YouTube-VIS2019, we tested only 794 new videos that were not present in the older version. In Table \ref{table:otherdataset}, the selected videos had a maximum ${T=36}$ and MBOR${\geq 0.25}$
.
As shown in the table, Eigen-Cluster VIS improved the mIoU by 0.66\% compared to MaskFreeVIS on the YouTube-VIS2021 dataset. On the OVIS dataset, the mIoU improvement reached 0.59\%, demonstrating the greater generalizability of the method.

\begin{table}[h!]
\centering
\caption{\bf{mIoU in Youtube-VIS 2021 and OVIS datasets.}}
\setlength{\tabcolsep}{8pt} 
\begin{tabular}{ccc}
\toprule
\textbf{model} & \textbf{Youtube-VIS 2021} & \textbf{OVIS} \\
\midrule
MaskfreeVIS & 57.53 &  52.34 \\
Eigen-Cluster VIS & \textbf{58.19} & \textbf{52.93} \\
\bottomrule
\end{tabular}
\label{table:otherdataset}
\end{table}

From Table \ref{table:otherdataset}, the graphs in Fig. \ref{fig:ablation23}  can be derived. In this chart, mIoU is reported based on three characteristics: occlusion, the number of instances, and the number of keypoints. Occlusion is measured using the MBOR criterion, and keypoints are extracted using the SIFT algorithm \cite{lowe2004distinctive}. Due to the unique characteristics of each dataset, we calculate the median value for each dataset and characteristic and then assess the accuracy on either side of these median values.

Fig. \ref{fig:ablation23}-a shows the relationship between MBOR values and segmentation performance. Generally, higher MBOR values lead to lower segmentation performance. Our proposed method improves accuracy by 0.93\% over MaskFreeVIS in the higher MBOR interval of the YouTube-VIS 2021 dataset, and by 1.19\% in the OVIS dataset.

Fig. \ref{fig:ablation23}-b  demonstrates how segmentation performance decreases as the number of instances increases. Our method outperforms MaskFreeVIS by 1.3\% in the higher instance interval of the YouTube-VIS 2021 dataset and by 0.62\% in the OVIS dataset.

Fig. \ref{fig:ablation23}-c  highlights the effect of the number of video foreground keypoints on segmentation accuracy. Fewer keypoints generally result in better detection due to smoother textures in the foreground. Our method shows a 1.17\% improvement over MaskFreeVIS in the higher keypoint interval of the YouTube-VIS 2021 dataset, and a 0.89\% improvement in the OVIS dataset.

As shown in the graph in Fig. \ref{fig:ablation23}, the proposed method faces challenges in scenarios with high occlusion. The weakly supervised approach struggles to effectively distinguish the features of different instances due to the presence of partial masks in cases involving significant occlusion and fast motion. Future work could address these limitations by leveraging video side information, such as optical flow, and exploring emerging techniques for performing optical flow in unsupervised manners \cite{hu2025mreiflow}. This could help mitigate these challenges and lead to further improvements.

\section{CONCLUSION}
\label{sec: conclusion}
In this paper, the Eigen-Cluster VIS model, which is a weakly supervised approach to video instance segmentation, was proposed. This method effectively narrowed the gap between the weakly supervised and fully supervised techniques without requiring ground truth masks at the video level. By introducing the Quality Cluster Coefficient module, a criterion for discriminating spatio-temporal instances without the need for ground truth regions was established. Additionally, the Temporal Eigenvalue Loss was introduced, which leverages the eigenvalue information of frames to create smoother transitions and enhance mask consistency across frames within a video clip. This simultaneous focus on both spatial and temporal dimensions, from the image patch level to the clip level, resulted in a VIS model that demonstrated strong performance compared to previous methods. Eigen-Cluster VIS offers potential for further development in both representation and optimization processes and can be integrated with fully unsupervised methods, which we plan to explore in our future work.


\bibliographystyle{IEEEtran}
\bibliography{main}

\vspace{11pt}

\vfill

\end{document}